\documentclass[acmtog,authorversion,nonacm]{acmart}
\acmSubmissionID{626}

\usepackage{booktabs} 

\usepackage{booktabs}
\usepackage{pifont} 
\newcommand{\xmark}{\ding{55}}
\usepackage{dblfloatfix}
\usepackage{enumitem}
\usepackage{multirow}
\usepackage{rotating}

\usepackage{colortbl}
\usepackage{xcolor}
\definecolor{Red}{RGB}{192, 0, 0}
\definecolor{Blue}{RGB}{12, 114, 186}
\definecolor{Yellow}{RGB}{218, 169, 20}
\definecolor{lightyellow}{RGB}{255,255,153}

\definecolor{HighlightBlue}{RGB}{0, 100, 148}
\definecolor{HighlightRed}{RGB}{230, 57, 70}

\definecolor{LightRed}{HTML}{ffe0e0}
\definecolor{LightBlue}{HTML}{def5ff}
\definecolor{LightYellow}{HTML}{FFF6DB}
\definecolor{LightGreen}{HTML}{eff9f0}

\usepackage{tikz}
\newcommand{\blackcircle}[1]{%
\tikz[baseline=(char.base),baseline=-0.7ex]{\node[shape=circle,fill=black,text=white,inner sep=0.5pt,font=\scriptsize] (char) {#1};}%
}

\newcommand{\OurMethod}{\emph{VideoPainter}}
\newcommand{\Benchmark}{\emph{VPBench}}
\newcommand{\TrainingData}{\emph{VPData}}

\usepackage[ruled]{algorithm2e} 

\SetAlFnt{\small}
\SetAlCapFnt{\small}
\SetAlCapNameFnt{\small}
\SetAlCapHSkip{0pt}

\acmJournal{TOG}

\begin{document}
\title{VideoPainter: Any-length Video Inpainting and Editing with Plug-and-Play Context Control}

\author{Yuxuan Bian}
\affiliation{
 \institution{The Chinese University of Hong Kong}
 \country{China}}
\email{yuxuanbian23@gmail.com}
\author{Zhaoyang Zhang}
\affiliation{
 \institution{Tencent ARC Lab}
 \country{China}}
\email{zhaoyangzhang@link.cuhk.edu.hk}
\author{Xuan Ju}
\affiliation{
 \institution{The Chinese University of Hong Kong}
 \country{China}}
\email{juxuan.27@gmail.com}
\author{Mingdeng Cao}
\affiliation{
 \institution{The University of Tokyo}
 \country{Japan}}
\email{mingdengcao@gmail.com}
\author{Liangbin Xie}
\affiliation{
 \institution{University of Macau}
 \country{China}}
\email{lb.xie@siat.ac.cn}
\author{Ying Shan}
\affiliation{
 \institution{Tencent ARC Lab}
 \country{China}}
\email{yingsshan@tencent.com}
\author{Qiang Xu}
\affiliation{
 \institution{The Chinese University of Hong Kong}
 \country{China}}
\email{qxu@cse.cuhk.edu.hk}
\renewcommand\shortauthors{Bian, Y. et al}

\begin{teaserfigure}
  \includegraphics[width=\textwidth]{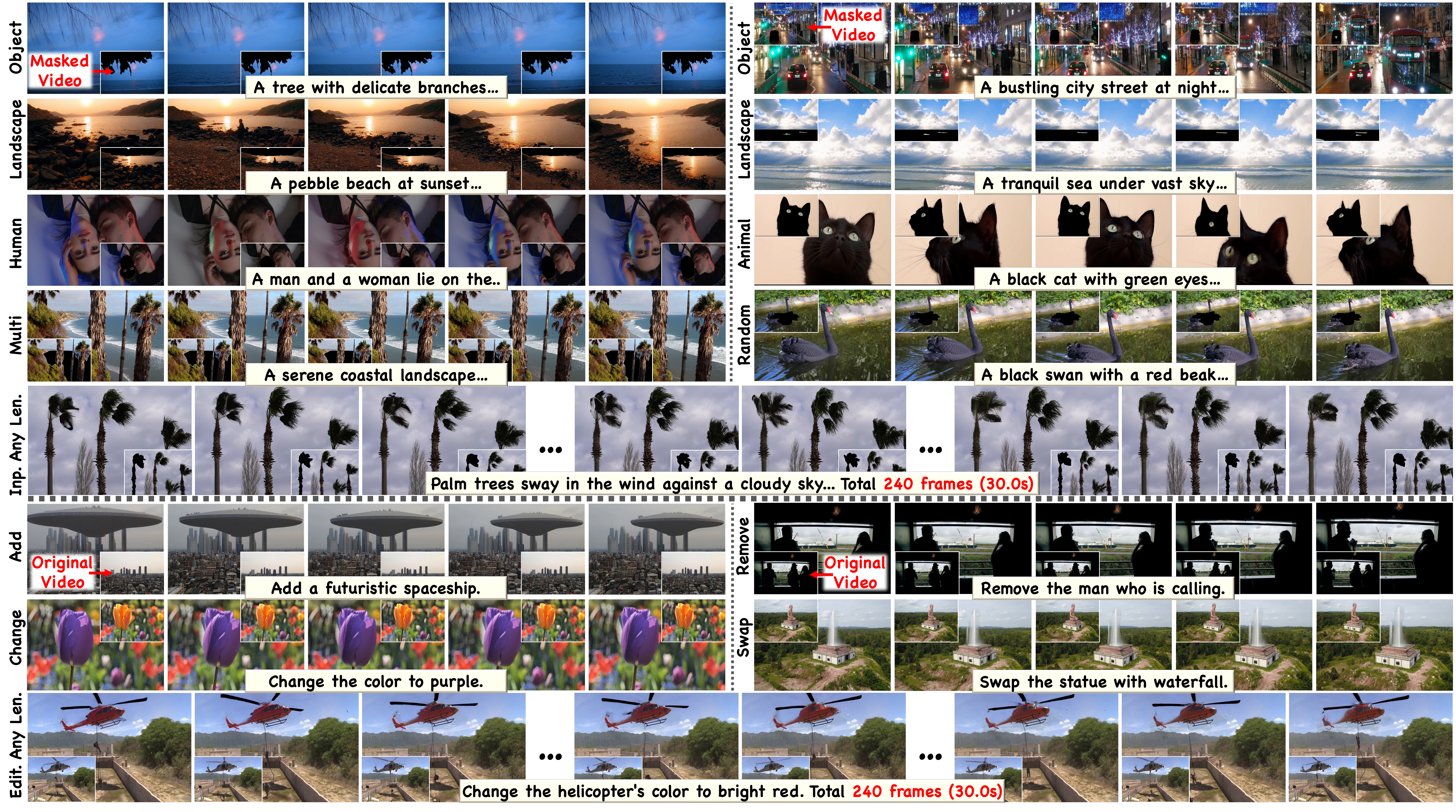}
  \vspace{-22pt}
  \caption{
  \textbf{\OurMethod~enables plug-and-play text-guided video inpainting and editing for any video length and pre-trained Diffusion Transformer with masked video and video caption~(user editing instruction).}
  The upper part demonstrates the effectiveness of \OurMethod~in various video inpainting scenarios, including object, landscape, human, animal, multi-region~(Multi), and random masks.
  The lower section demonstrates the performance of \OurMethod~in video editing, including adding, removing, changing attributes, and swapping objects.
  In both video inpainting and editing, we demonstrate strong ID consistency in generating long videos~(Any Len.).
  %
  \noindent \textbf{Project page for this paper is at: } \href{https://yxbian23.github.io/project/video-painter}{\textbf{https://yxbian23.github.io/project/video-painter}}
  }
  \Description{Teaser}
  \label{fig:teaser}
\end{teaserfigure}

\begin{abstract}
Video inpainting, crucial for the media industry, aims to restore corrupted content. 
However, current methods relying on limited pixel propagation or single-branch image inpainting architectures face challenges with generating fully masked objects, balancing background preservation with foreground generation, and maintaining ID consistency over long video.
To address these issues, we propose \OurMethod, an efficient dual-branch framework featuring a lightweight context encoder. This plug-and-play encoder processes masked videos and injects background guidance into any pre-trained video diffusion transformer, generalizing across arbitrary mask types, enhancing background integration and foreground generation, and enabling user-customized control.
We further introduce a strategy to resample inpainting regions for maintaining ID consistency in any-length video inpainting. 
Additionally, we develop a scalable dataset pipeline using advanced vision models and construct \TrainingData~and \Benchmark—the largest video inpainting dataset with segmentation masks and dense caption (>390K clips) —to support large-scale training and evaluation.
We also show \OurMethod's promising potential in downstream applications such as video editing.
Extensive experiments demonstrate \OurMethod's state-of-the-art performance in any-length video inpainting and editing across $8$ key metrics, including video quality, mask region preservation, and textual coherence.

\end{abstract}

%
%

%
%


\maketitle

\section{Introduction}\label{sec-1-intro}
Video inpainting~\cite{quan2024deep}, which aims to restore the corrupted video while
maintaining coherence, facilitates numerous applications, including try-on~\cite{fang2024vivid}, film production~\cite{polyak2024movie}, and video editing~\cite{sun2024diffusion}. Recently, Diffusion Transformers~(DiT)~\cite{peebles2023scalable,sora} have shown promise in video generation, leading to the exploration of generative video inpainting~\cite{zhang2024avid,zi2024cococo}.

\begin{figure*}[htbp]
  \centering
  \includegraphics[width=0.98\textwidth]{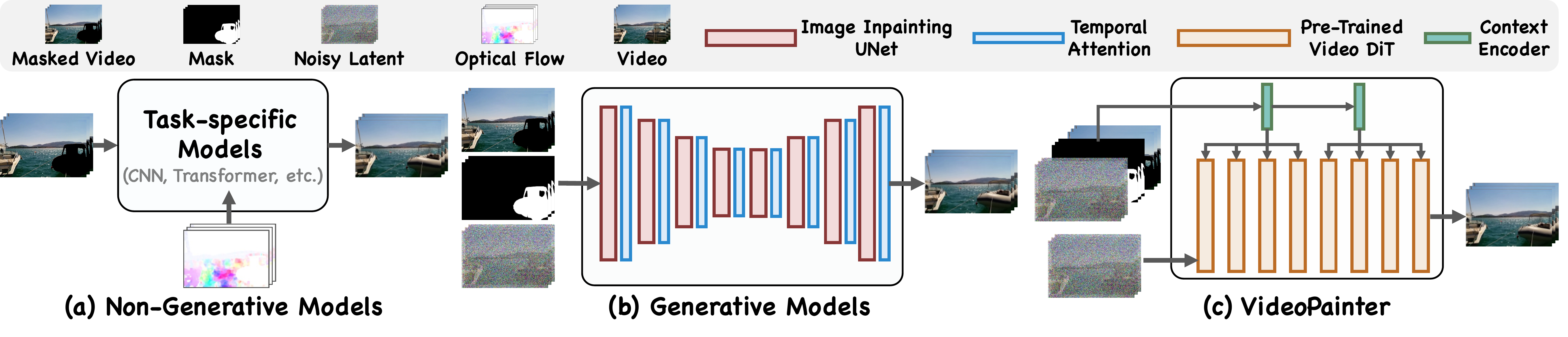} 
  \vspace{-4pt}
  \caption{Framework Comparison. 
  Non-generative approaches, limited to pixel propagation from backgrounds, fail to inpaint fully segmentation-masked objects. Generative methods adapt single-branch image inpainting models to video by adding temporal attention, struggling to maintain background fidelity and generate foreground contents in one model. In contrast, \OurMethod~implements a dual-branch architecture that leverages an efficient context encoder with any pre-trained DiT, decoupling video inpainting to background preservation and foreground generation, and enabling plug-and-play video inpainting control.
  }
  \vspace{-4pt}
  \label{fig:framework}
\end{figure*}

Existing approaches, as illustrated in Fig. \ref{fig:framework}, can be broadly categorized into two types: (1) Non-Generative methods~\cite{Propainter,EEFVI,transformer1} depend on limited pixel feature propagation~(physical constraints or model architectural priors), which only take masked videos as inputs and cannot generate fully segmentation-masked objects. (2) Generative methods~\cite{zhang2024avid,zi2024cococo,wang2024videocomposer} extend single-branch image inpainting architectures~\cite{sd2inpainting} to video by incorporating temporal attention, which struggles to balance background preservation and foreground generation in one model and obtain inferior temporal coherence compared to native video DiTs.
Moreover, both paradigms neglect long video inpainting and struggle to maintain consistent object identity with long videos.

This motivates us to decompose video inpainting into background preservation and foreground generation and adopt a dual-branch architecture in DiTs, where we can incorporate a dedicated context encoder for masked video feature extraction while utilizing the pre-trained DiT's capabilities to generate semantic coherent video content conditioned on both the preserved background and text prompts.
Similar observations have been made in image inpainting research, notably in BrushNet~\cite{ju2024brushnet} and ControlNet~\cite{zhang2023adding}. 
However, directly applying their architecture to video DiTs presents several challenges: (1) Given Video DiT's robust generative foundation and heavy model size, replicating the full/half-giant Video DiT backbone as the context encoder would be unnecessary and computationally prohibitive. (2) Unlike BrushNet's pure convolutional control branch, DiT's tokens in masked regions inherently contain background information due to global attention, complicating the distinction between masked and unmasked regions in DiT backbones.  (3) ControlNet lacks feature injection across all layers, hindering dense background control for inpainting tasks.

To address these challenges, we introduce \OurMethod, which enhances pre-trained DiT with a lightweight context encoder comprising only 6\% of the backbone parameters, to form the first efficient dual-branch video inpainting architecture. \OurMethod~features three main components: (1) A streamlined context encoder with just two layers, which integrates context features into the pre-trained DiT in a group-wise manner, ensuring efficient and dense background guidance. (2) Mask-selective feature integration to clearly distinguish the tokens of the masked and unmasked region. 
(3) A novel inpainting region ID resampling technique to efficiently process videos of any length while maintaining ID coherence. By freezing the pre-trained context encoder and DiT backbone, and adding an ID-Adapter, we enhance the backbone's attention sampling by concatenating the original key-value vectors with the inpainting region tokens. During inference, inpainting region tokens from previous clips are appended to the current key-value vectors, ensuring the long-term preservation of target IDs.
Notably, our \OurMethod~supports plug-and-play and user-customized control.

For large-scale training, we develop a scalable dataset pipeline using advanced vision models~\cite{gpt4,ravi2024sam,zhang2024recognize}, constructing the largest video inpainting dataset, \TrainingData, and benchmark, \Benchmark, with over 390K clips featuring precise segmentation masks and dense text captions. We further demonstrate \OurMethod's potential by establishing an inpainting-based video editing pipeline that delivers promising results.

To validate our approach, we compare \OurMethod~against previous state-of-the-art~(SOTA) baselines and a single-branch fine-tuning setup that combines noisy latent, masked video latent, and mask at the input channel. \OurMethod~demonstrates superior performance in both training efficiency and final results.

In summary, our contributions are as follows:
\vspace{-2pt}
\begin{itemize}[leftmargin=*]
    \setlength\itemsep{0.1em}
     \item We propose \textbf{\textit{\OurMethod}}, the first dual-branch video inpainting framework that supports plug-and-play background controls.

     \item We design a lightweight context encoder for efficient and dense background control, and inpainting region ID resampling for ID consistency in any-length video inpainting and editing.
     
    \item We introduce \textbf{\textit{\TrainingData}}, the largest video inpainting datasets comprising over 390K clips ($\textgreater 866.7$ hours), and \textbf{\textit{\Benchmark}}, both featuring precise masks and detailed video captions.

    \item Experiments show \OurMethod~achieves state-of-the-art performance across $8$ metrics including video quality, masked region preservation, and text alignment in video inpainting and editing.

\end{itemize}

\section{Related Work}\label{sec-2-related_Work}

\subsection{Video Inpainting}
Video inpainting approaches can be broadly classified into two categories based on whether they possess generative capabilities:
%
%
\paragraph{Non-generative methods.} 
These methods~\cite{Propainter,EEFVI,3dcnn3,transformer4,optical7} leverage architecture priors to facilitate pixel propagation. This includes utilizing local perception of 3D CNNs~\cite{3dcnn1,3dcnn2,3dcnn3,temporal_shift}, and exploiting the global perception of attention to retrieve and aggregate tokens with similar texture for filling masked video~\cite{transformer1,transformer2,transformer3,transformer4}.
They also introduce various physical quantities, especially optical flow, as auxiliary conditions as it simplifies RGB pixel inpainting by completing less complex flow fields~\cite{optical1,optical2,optical3,optical4,optical5,optical6,optical7}. 
However, they are only effective for partial object occlusions with random masks but face significant limitations when inpaint fully masked regions due to insufficient contexts.
%
%
\paragraph{Generative methods.}
Recent advances in generative foundation models~\cite{rombach2022high,guo2023animatediff} have sparked numerous approaches that leverage additional modules or training strategies to extend backbones' capabilities for video inpainting~\cite{zhang2024avid,zi2024cococo,wang2024videocomposer}.
%
%
AVID~\cite{zhang2024avid} and COCOCO~\cite{zi2024cococo} represent the most related recent works. Both adopt a similar implementation by augmenting Stable Diffusion Inpainting~\cite{sd2inpainting} with trainable temporal attention layers. This architecture includes per-frame region filling based on the image inpainting backbone and temporal smoothing with temporal attention. 
Despite showing promising results for both random and segmentation masks due to their generative abilities, they struggle to balance background preservation and foreground generation with text caption~\cite{ju2024brushnet,li2024brushedit} within the single backbone.
AVID also explores any-length video inpainting by smoothing latent at segment boundaries and using the middle frame as the ID reference. 
%
In contrast, \OurMethod~is a dual-branch framework by decoupling video inpainting into foreground generation and background-guided preservation. It employs an efficient context encoder to guide any pre-trained DiT, facilitating plug-and-play control.
Furthermore, \OurMethod~also introduces a novel inpainting region ID resampling technique that enables ID consistency in any-length video inpainting.


\begin{table}[t]
\centering
\caption{Comparison of video inpainting datasets. Our \TrainingData~is the largest video inpainting dataset to date, comprising over 390K
high-quality clips with segmentation masks, video captions, and masked region descriptions.}
\vspace{-6pt}
\label{tab-dataset_comparison}
\setlength{\tabcolsep}{0.9mm}
\scalebox{0.71}{
\begin{tabular}{l|c|c|c|c}
\toprule
Dataset & \#Clips & Duration & Video Caption & Masked Region Desc. \\
\midrule
DAVIS~\cite{perazzi2016benchmark} & 0.4K & 0.1h & \xmark & \xmark \\
YouTube-VOS~\cite{xu2018youtube} & 4.5K & 5.6h & \xmark & \xmark \\
VOST~\cite{tokmakov2023breaking} & 1.5K & 4.2h & \xmark & \xmark \\
MOSE~\cite{ding2023mose} & 5.2K & 7.4h & \xmark & \xmark \\
LVOS~\cite{hong2023lvos} & 1.0K & 18.9h & \xmark & \xmark \\
SA-V~\cite{ravi2024sam} & 642.6K & 196.0h & \xmark & \xmark \\
\bottomrule
Ours & 390.3K & 866.7h & \checkmark & \checkmark \\
\bottomrule
\end{tabular}
}
\vspace{-0.4cm}
\end{table}

\subsection{Video Inpainting Datasets}
Recent advances in segmentation~\cite{ravi2024sam} have created many video segmentation datasets~\cite{perazzi2016benchmark,xu2018youtube,hong2023lvos,ding2023mose,tokmakov2023breaking,VISOR2022}. Among these, DAVIS~\cite{perazzi2016benchmark} and YouTube-VOS~\cite{xu2018youtube} have become prominent benchmarks for video inpainting due to their high-quality masks and diverse object categories. However, the existing datasets face two primary limitations: (1) insufficient scale to meet the data requirements of generative models, and (2) the absence of crucial control conditions necessary for generating masked objects such as video captions. 
%
%
In contrast, as shown in Tab. \ref{tab-dataset_comparison}, we developed a scalable dataset pipeline based on state-of-the-art vision understanding models~\cite{zhang2024recognize,ravi2024sam,gpt4}, and constructed the largest video inpainting dataset to date with over 390K clips, each annotated with segmentation masks and dense video captions.

\begin{figure}[t]
  \centering
  \includegraphics[width=0.48\textwidth]{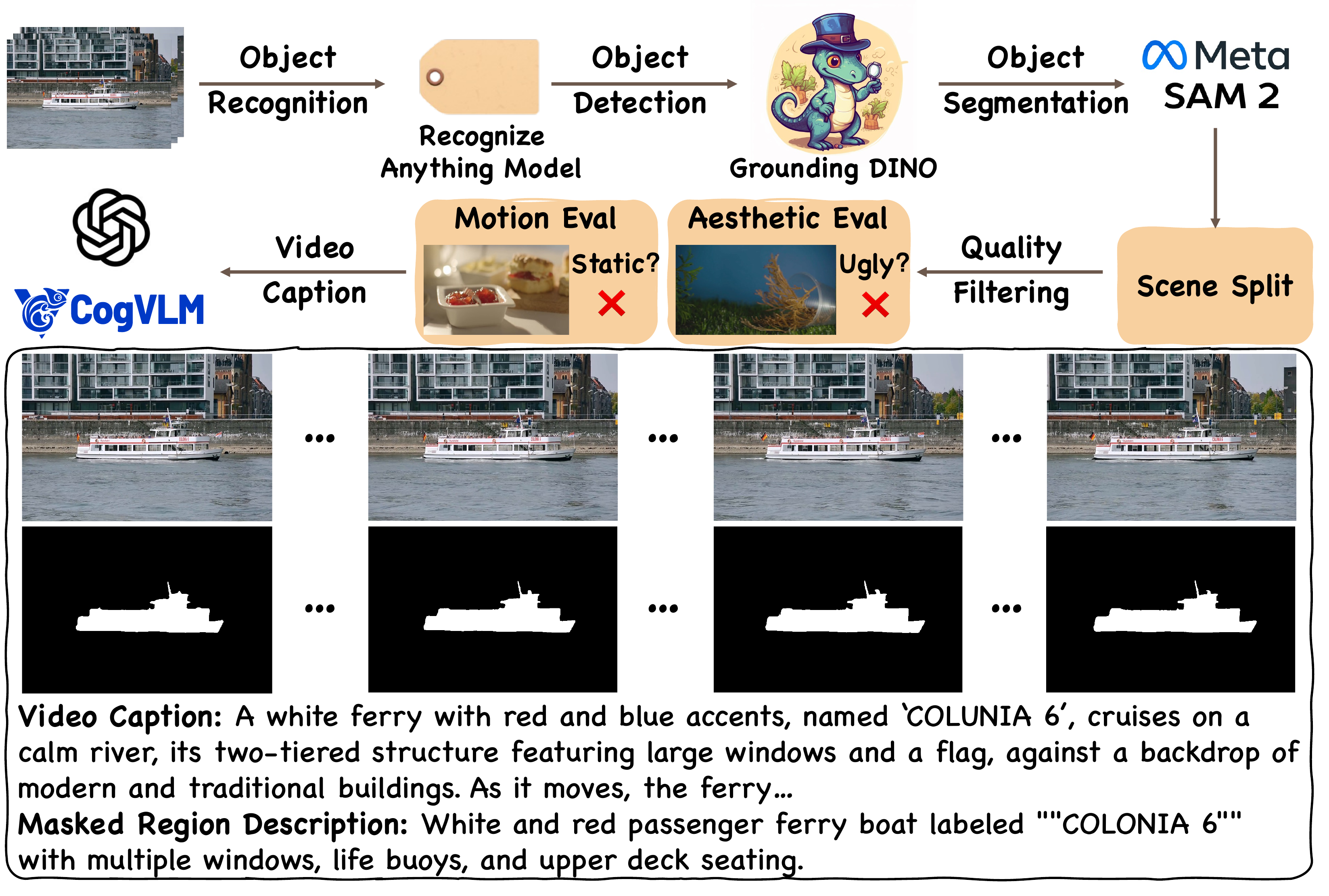} 
  \vspace{-20pt}
  \caption{Dataset Construction Pipeline. It consists of five pre-processing steps: collection, annotation, splitting, selection, and captioning.}
  \vspace{-14pt}
  \label{fig:dataset}
\end{figure}
\section{Method}\label{sec-3-method}

\begin{figure*}[htbp]
  \centering
  \includegraphics[width=0.95\textwidth]{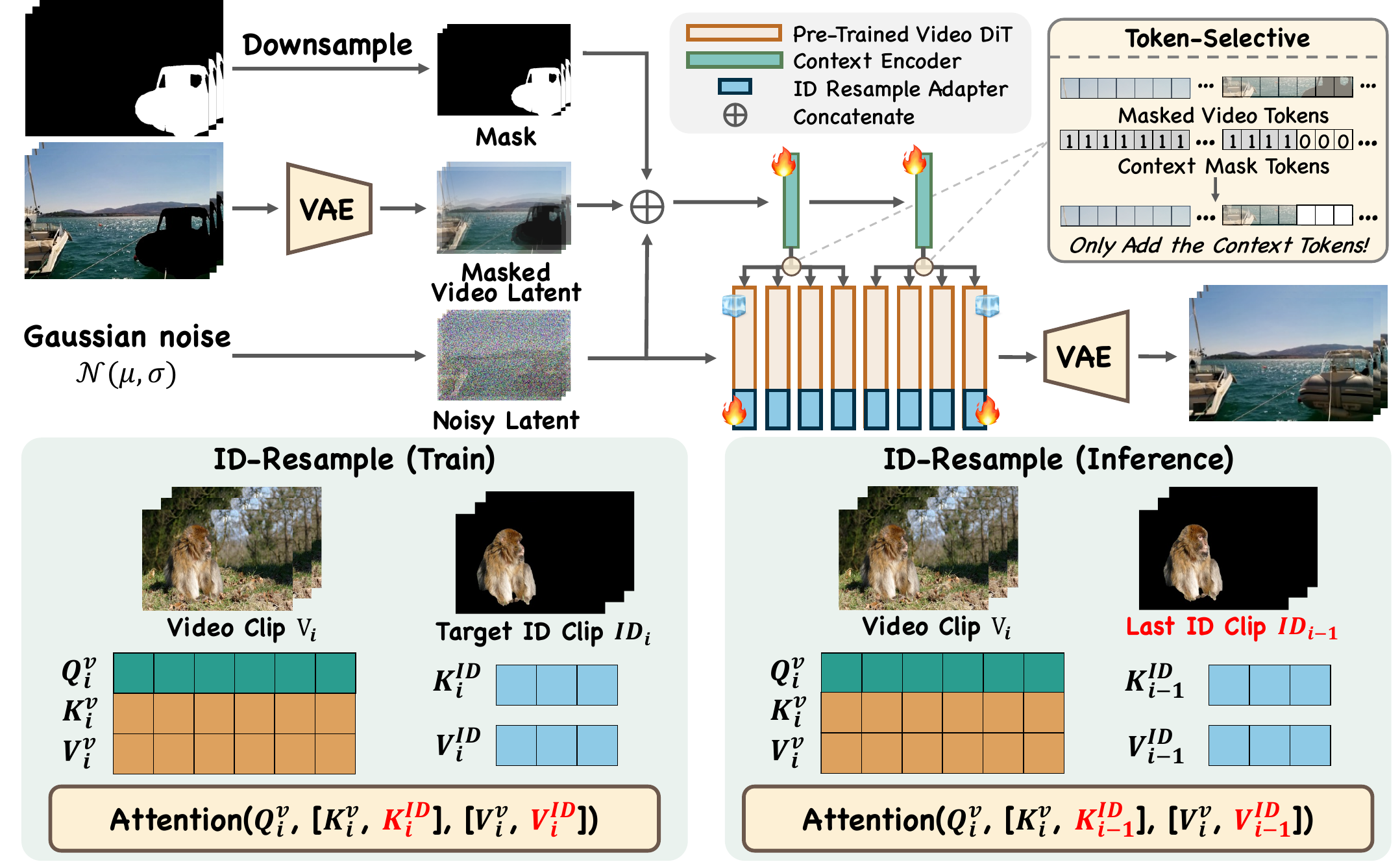}
  \vspace{-6pt}
  \caption{
   \textbf{Model overview.} 
   \textbf{The upper figure} shows the architecture of \OurMethod. The context encoder performs video inpainting based on concatenation of the noisy latent, downsampled masks, and masked video latent via VAE. Features extracted by the context encoder are integrated into the pre-trained DiT in a group-wise and token-selective manner, where two encoder layers modulate the first and second halves of the DiT, respectively, and only the background tokens will be integrated into the backbone to prevent information ambiguity.
   %
   %
   \textbf{The lower figure} illustrates the inpainting ID region resampling with the ID Resample Adapter. During training, tokens of the current masked region are concatenated to the KV vectors, enhancing ID preservation of the inpainting region. During inference, the ID tokens of the last clip are concatenated to the current KV vectors, maintaining ID consistency with the last clip by resampling.
  }
  \vspace{-8pt}
  \label{fig:method}
\end{figure*}

Sec. \ref{sec-3-1} and Fig. \ref{fig:dataset} illustrate our pipeline for building \TrainingData~and \Benchmark. Sec. \ref{sec-3-2} and Fig. \ref{fig:method} show our dual-branch \OurMethod. Sec. \ref{sec-3-3} and Sec. \ref{sec-3-4} introduce our inpainting region ID resampling approach for any-length video inpainting and plug-and-play control.

\subsection{\TrainingData~and \Benchmark~Construction Pipeline}\label{sec-3-1}

To address the challenges of limited size and lack of text annotations, we present a scalable dataset pipeline leveraging advanced vision models~\cite{ravi2024sam,gpt4,zhang2024recognize}. This leads to \TrainingData~and \Benchmark, the largest video inpainting dataset and benchmark with precise masks and video/masked region captions.
As shown in Fig. \ref{fig:dataset}, the pipeline involves $5$ steps: collection, annotation, splitting, selection, and captioning.

\noindent \textbf{Collection.} We chose Videvo and Pexels~\footnote{Videvo: \url{https://www.videvo.net/}, Pexels: \url{https://www.pexels.com/}} as our data sources. We finally obtained around $450K$ videos from these sources.

\noindent \textbf{Annotation.} For each collected video, we implement a cascaded workflow for automated annotation:
\begin{itemize}[leftmargin=*,itemsep=-0.1em]
    \vspace{-4pt}
    \item[\ding{224}] We employ the Recognize Anything Model~\cite{zhang2024recognize} for open-set video tagging to identify primary objects.
    \item[\ding{224}] Based on the detected object tags, we utilize Grounding DINO~\cite{liu2023grounding} to detect bounding boxes for objects at fixed intervals.
    \item[\ding{224}] These bounding boxes serve as prompts for SAM2~\cite{ravi2024sam}, which generates high-quality mask segmentations.
    \item[\ding{224}] Then we employ rigorous filtering criteria: inter-frame mask area variation \( \Delta <20\% \) and frame coverage maintained between $30\%-70\%$ to ensure reliable segmentation masks quality.
\end{itemize}

\noindent \textbf{Splitting.} Scene transitions may occur while tracking the same object from different angles, causing disruptive view changes. We utilize PySceneDetect~\cite{castellano2024pyscenedetect} to identify scene transitions and subsequently partition the masks. 
Then we segmented the sequences into 10-second intervals and discarded short clips (< 6s).

\noindent \textbf{Selection.} We employ $3$ key criteria: (1) Aesthetic Quality, evaluated using the Laion-Aesthetic Score Predictor~\cite{laion5b}; (2) Motion Strength, predicted by optical flow measurements using the RAFT\cite{raft}; and (3) Content Safety, assessed via the Stable Diffusion Safety Checker~\cite{rombach2022high}. 

\noindent \textbf{Captioning.} As Tab. \ref{tab-dataset_comparison} shows, existing video segmentation datasets lack textual annotations, primary conditions in generation~\cite{pixart,dalle3}, creating a data bottleneck for applying generative models to video inpainting. Therefore, we leverage SOTA VLMs, specifically CogVLM2~\cite{wang2023cogvlm} and GPT-4o~\cite{gpt4}, to uniformly sample keyframes and generate dense video captions and detailed descriptions of the masked objects. 

\subsection{Dual-branch Inpainting Control}\label{sec-3-2}

We incorporate masked video features into the pre-trained diffusion transformer~(DiT) via an efficient context encoder, to decouple the background context extraction and foreground generation. This encoder processes a concatenated input of noisy latent, masked video latent, and downsampled masks.
Specifically, the noisy latent provides information about the current generation. The masked video latent, extracted via VAE, aligns with the pre-trained DiT's latent distribution. We apply cubic interpolation to downsample masks, ensuring dimensional compatibility between masks and latents.

Based on DiT's inherent generative abilities~\cite{sora}, the control branch only needs to extract contextual cues to guide the backbone in preserving background and generating foreground. Therefore, instead of previous heavy approaches that duplicate half or all of the backbone~\cite{ju2024brushnet,zhang2023adding}, \OurMethod~employs a lightweight design by cloning only the first two layers of pre-trained DiT, accounting for merely 6\% of the backbone parameters. The pre-trained DiT weights provide a robust prior for extracting masked video features.
The context encoder features are integrated into the frozen DiT in a group-wise, token-selective manner. 
The group-wise feature integration is formulated as follows: the first layer's features are added back to the initial half of the backbone, while the second layer's features are integrated into the latter half, achieving lightweight and efficient context control. 
The token-selective mechanism is a pre-filtering process, where only tokens representing pure background are added back, while others are excluded from integration, as shown in the upper right of Fig. \ref{fig:method}. This ensures that only the background context is fused into the backbone, preventing potential ambiguity during backbone generation.
%

%
The feature integration is shown in Eq.~\ref{eq:insertion}. $\epsilon _{\theta}\left( z_t,t,C \right) _i$ indicates the feature of the $i$-th layer in DiT $\epsilon _{\theta}$ with $i\sim \left[ 1,n \right]$, where $n$ is the number of layers. The same notation applies to $\epsilon _{\theta}^{VideoPainter}$, which takes the concatenated noisy latent $z_t$, masked video latent $z_{0}^{masked}$, and downsampled mask $m^{resized}$ as input. The concatenation operation is denoted as $\left[ \cdot \right]$. $\mathcal{Z}$ is the zero linear operation. 
\begin{equation}
\scalebox{0.78}{$\epsilon _{\theta}\left( z_t,t,C \right) _i=\epsilon _{\theta}\left( z_t,t,C \right) _i+\mathcal{Z} \left( \epsilon _{\theta}^{VideoPainter}\left( \left[ z_t,z_{0}^{masked},m^{resized} \right] ,t \right) _{i//\frac{n}{2}} \right)$}
\label{eq:insertion}
\end{equation}

\subsection{Target Region ID Resampling}\label{sec-3-3}

While current DiTs show promise in handling temporal dynamics~\cite{kling,bian2024multi}, they struggle to maintain smooth transitions and long-term identity consistency.

\noindent \emph{\textbf{Smooth Transition.}} Following AVID~\cite{zhang2024avid}, we employ overlapping generation and weighted average to maintain consistent transitions. Additionally, we utilize the last frame of the previous clip (before overlap) as the first frame of the current clip's overlapping region to ensure visual appearance continuity.

\noindent \emph{\textbf{Identity Consistency.}} To maintain identity consistency in the long video, we introduce an inpainting region ID resampling method, as shown in lower Fig. \ref{fig:method}.
During training, we freeze both the DiT and the context encoder. Then we add trainable ID-Resample Adapters into the frozen DiT~(LoRA), enabling ID resampling functionality. Specifically, tokens from the current masked region, which contain the desired ID, are concatenated with the KV vectors, thereby enhancing ID preservation in the inpainting region through additional KV resampling.
During inference, we prioritize maintaining ID consistency with the inpainting region tokens from the previous clip, as it represents the most temporally proximate generated result.
%
Specifically, given current \(Q^v_i\), \(K^v_i\), and \(V^v_i\), we concatenate tokens containing ID information (\(K^{id}_i\) and \(V^{id}_i\)) to the current KV pairs (during training, these are tokens from the current inpainting region; during inference, from the previous clip's inpainting region). This forms new KV-vectors \([K^v_i, K^{id}_i]\) and \([V^v_i, V^{id}_i]\) (where \([\cdot,\cdot]\) denotes concatenation), enabling the model to sample necessary ID information and better maintain ID consistency.

\subsection{Plug-and-Play Control}\label{sec-3-4}

Our plug-and-play framework demonstrates versatility across two aspects: it supports various stylization backbones or LoRAs and is compatible with both text-to-video (T2V)~\cite{yang2024cogvideox,nvidia_cosmos_2025} and image-to-video (I2V)~\cite{guo2024i2v,shi2024motion} DiT architectures. The I2V compatibility particularly enables seamless integration with existing image inpainting capabilities. When utilizing an I2V DiT backbone, \OurMethod~requires only one additional step: generating the initial frame using any image inpainting model guided by the masked region's text caption. 
This inpainted frame then serves as both the image condition and the first masked video frame. These capabilities further demonstrate the exceptional transferability and versatility of \OurMethod.
\section{Experiments}\label{sec-4-exp}

\subsection{Implementation details}
\OurMethod~is built upon a pre-trained Image-to-Video Diffusion Transformer CogVideo-5B-I2V~\cite{yang2024cogvideox} (by default) and its Text-to-Video version. 
In training, we use \TrainingData~at a $480\times720$ resolution, learning rate $1 \times 10^{-5}$, batch size  $1$ for both the context encoder~($80,000$ steps) and the ID Resample Adapter~($2,000$ steps) in two stages with AdamW. 
In training, we randomly sample dilation and erosion with kernel sizes $\in [8, 32]$ to enhance robustness to mask precision. This also enables our random-mask inpainting.

\paragraph{\textbf{Benchmarks.}}
In video inpainting, we employ Davis~\cite{perazzi2016benchmark} as the benchmark for random masks and \Benchmark~for segmentation-based masks. \Benchmark~consists of $100$ 6-second videos for standard video inpainting, and $16$ videos with an average duration of more than $30$ seconds for long video inpainting. The \Benchmark~includes diverse content including objects, humans, animals, landscapes, and multi-range masks.
For video editing evaluation, we also utilize \Benchmark, which includes four fundamental editing operations (add, remove, swap, and change) and comprises $45$ 6-second videos and $9$ videos with an average duration of $30$ seconds.

\paragraph{\textbf{Metrics.}} We consider $8$ metrics from three aspects: masked region preservation, text alignment, and video generation quality.
\begin{itemize}[leftmargin=*,itemsep=-0.1em]
    \item[$\bullet$] \textit{Masked Region Preservation.} We follow previous works using standard PSNR~\cite{psnr}, LPIPS~\cite{lpips}, SSIM~\citep{ssim}, MSE~\cite{mse} and MAE~\cite{mae} in the unmasked region among the generated video and the original video.

    \item[$\bullet$] \textit{Text Alignment.} We employ CLIP Similarity (CLIP Sim)~\cite{clipsim} to assess the semantic consistency between the generated video and its corresponding text caption. We also measure CLIP Similarity within the masked regions (CLIP Sim (M)).

    \item[$\bullet$] \textit{Video Generation Quality.} 
    Following previous methods, we use FVID~\cite{fvid} to measure the generated video quality.
\end{itemize}

\subsection{Video Inpainting}
\begin{figure*}[htbp]
  \centering
  \includegraphics[width=0.95\textwidth]{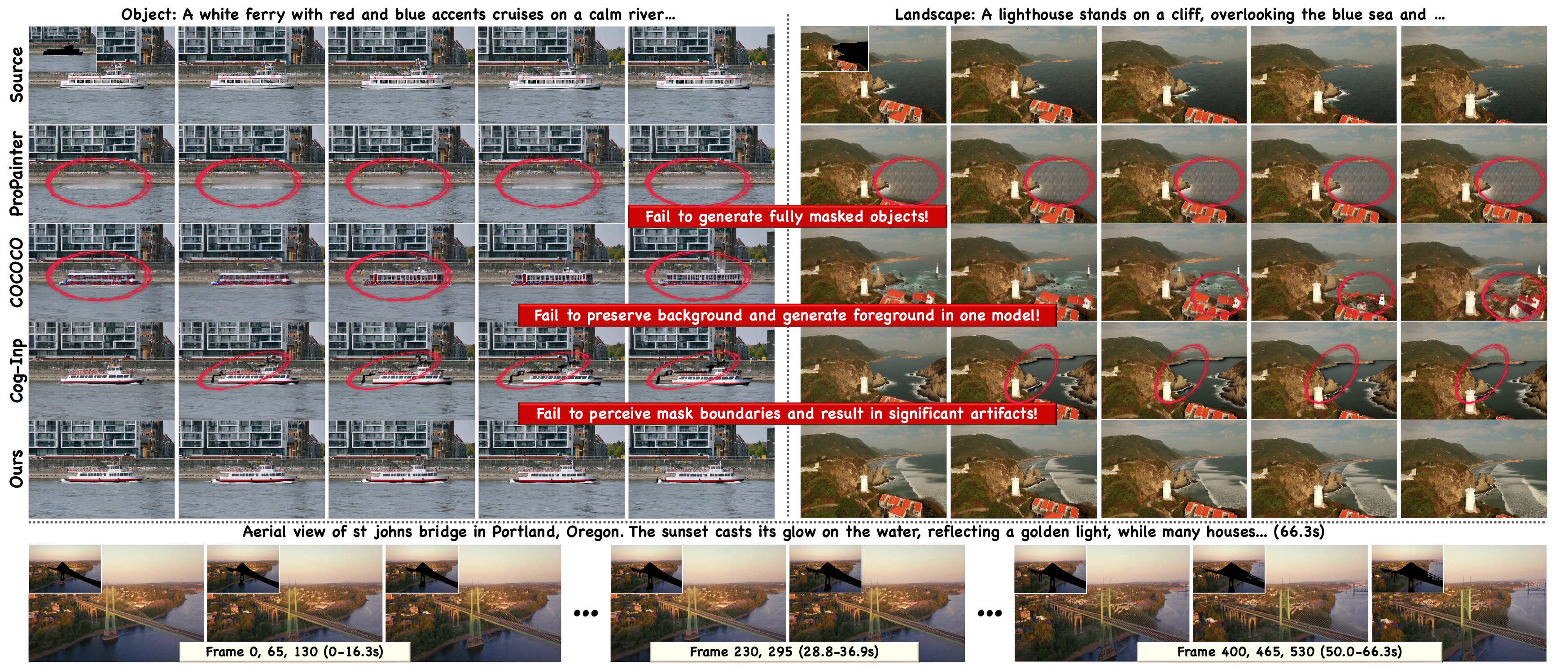} 
  \caption{Comparison of previous inpainting methods and \OurMethod~on standard and long video inpainting. More visualizations are in the demo video.}
  \label{fig:inp}
\end{figure*}

\paragraph{Quantitative comparisons}

Tab.~\ref{tab:inp} shows the quantitative comparison on \Benchmark~and Davis~\cite{perazzi2016benchmark}. 
We compare the inpainting results of non-generative ProPainter~\cite{Propainter}, generative COCOCO~\cite{zi2024cococo}, and Cog-Inp~\cite{yang2024cogvideox}, a strong baseline proposed by us, which inpaint first frame using image inpainting models and use the I2V backbone to propagate results with the latent blending operation~\cite{avrahami2023blended}.
In the segmentation-based \Benchmark, ProPainter, and COCOCO exhibit the worst performance across most metrics, primarily due to the inability to inpaint fully masked objects and the single-backbone architecture's difficulty in balancing the competing background preservation and foreground generation, respectively.
In the random mask benchmark Davis, ProPainter shows improvement by leveraging partial background information. 
However, \OurMethod~achieves optimal performance across segmentation (standard and long length) and random masks through its dual-branch architecture that effectively decouples background preservation and foreground generation.
%

\paragraph{Qualitative comparisons}

The qualitative comparison with previous video inpainting methods is shown in Fig.~\ref{fig:inp}. 
\OurMethod~consistently shows exceptional results in video coherence, quality, and alignment with text caption. Notably, ProPainter fails to generate fully masked objects because it only depends on background pixel propagation instead of generating. 
While COCOCO demonstrates basic functionality, it fails to maintain consistent ID in inpainted regions ( inconsistent vessel appearances and abrupt terrain changes) due to its single-backbone architecture attempting to balance background preservation and foreground generation.
Cog-Inp achieves basic inpainting results; however, its blending operation's inability to detect mask boundaries leads to significant artifacts.
Moreover, \OurMethod~can generate coherent videos exceeding one minute while maintaining ID consistency through our ID resampling.

\begin{table}[t]
\centering
\scriptsize
\caption{\textbf{Quantitative comparisons among \OurMethod~and other video inpainting models in \Benchmark~for segmentation mask~(Standard~(S) and Long~(L) Video) and Davis for random mask}: ProPainter~\cite{Propainter}, COCOCO~\cite{zi2024cococo}, and Cog-Inp~\cite{yang2024cogvideox}. Metrics include masked region preservation, text alignment, and video quality. \textcolor{Red}{\textbf{Red}} stands for the best, \textcolor{Blue}{\textbf{Blue}} stands for the second best. }
\scalebox{0.85}{
\setlength{\tabcolsep}{0.2mm}{
\begin{tabular}{cl|ccccc|cc|c}
\toprule
\multicolumn{2}{c|}{\textbf{Metrics}}  & \multicolumn{5}{c|}{\textbf{Masked Region Preservation}} & \multicolumn{2}{c|}{\textbf{Text Alignment}} & \textbf{Video Quality} \\
\midrule
\multicolumn{2}{c|}{\textbf{Models}}    & \textbf{PSNR}$\uparrow$   & \textbf{SSIM}$\uparrow$  & \textbf{LPIPS}$_{^{\times 10^2}}$$\downarrow$   & \textbf{MSE}$_{^{\times 10^2}}$$\downarrow$     & \textbf{MAE}$_{^{\times 10^2}}$$\downarrow$  & \textbf{CLIP Sim}$\uparrow$      & \textbf{CLIP Sim (M)}$\uparrow$  & \textbf{FVID}$\downarrow$   \\ \midrule
\multirow{4}{*}{\rotatebox{90}{\textbf{\Benchmark-S}}}  & 
\textbf{\xspace ProPainter} & 20.97 & \textcolor{Blue}{\textbf{0.87}} & 
9.89 & 1.24 & \textcolor{Blue}{\textbf{3.56}} & 7.31 & 17.18 & 0.44  \\  &
\textbf{\xspace COCOCO} & 19.27 & 0.67 & 
14.80 & 1.62 & 6.38 & 7.95 & 20.03 & 0.69  \\  &
\textbf{\xspace Cog-Inp} & \textcolor{Blue}{\textbf{22.15}} & 0.82 & 
\textcolor{Blue}{\textbf{9.56}} & \textcolor{Blue}{\textbf{0.88}} & 3.92 & \textcolor{Blue}{\textbf{8.41}} & \textcolor{Blue}{\textbf{21.27}} & \textcolor{Blue}{\textbf{0.18}}  \\  &
\textbf{\xspace Ours} & \textcolor{Red}{\textbf{23.32}}&\textcolor{Red}{\textbf{0.89}}&\textcolor{Red}{\textbf{6.85}}&\textcolor{Red}{\textbf{0.82}}&\textcolor{Red}{\textbf{2.62}}&\textcolor{Red}{\textbf{8.66}}&\textcolor{Red}{\textbf{21.49}} & \textcolor{Red}{\textbf{0.15}} \\ 
\midrule
\multirow{4}{*}{\rotatebox{90}{\textbf{\Benchmark-L}}} &  
\textbf{\xspace ProPainter} & \textcolor{Blue}{\textbf{20.11}} & \textcolor{Blue}{\textbf{0.84}} & 
\textcolor{Blue}{\textbf{11.18}} & \textcolor{Blue}{\textbf{1.17}} & \textcolor{Blue}{\textbf{3.71}} & 9.44 & 17.68 & 0.48  \\  &
\textbf{\xspace COCOCO} & 19.51 & 0.66 & 
16.17 & 1.29 & 6.02 & 11.00 & 20.42 & 0.62  \\  &
\textbf{\xspace Cog-Inp} & 19.78 & 0.73 & 
12.53 & 1.33 & 5.13 & \textcolor{Blue}{\textbf{11.47}} & \textcolor{Blue}{\textbf{21.22}} & \textcolor{Blue}{\textbf{0.21}}  \\  &
\textbf{\xspace Ours} & \textcolor{Red}{\textbf{22.19}}&\textcolor{Red}{\textbf{0.85}}&\textcolor{Red}{\textbf{9.14}}&\textcolor{Red}{\textbf{0.71}}&\textcolor{Red}{\textbf{2.92}}&\textcolor{Red}{\textbf{11.52}}&\textcolor{Red}{\textbf{21.54}} &\textcolor{Red}{\textbf{0.17}}   \\
\midrule
\multirow{4}{*}{\rotatebox{90}{\textbf{Davis}}} &  

\textbf{\xspace ProPainter} & \textcolor{Blue}{\textbf{23.99}} & \textcolor{Blue}{\textbf{0.92}} & 
\textcolor{Blue}{\textbf{5.86}} & 0.98 & \textcolor{Blue}{\textbf{2.48}} & \textcolor{Red}{\textbf{7.54}} & 16.69 & \textcolor{Blue}{\textbf{0.12}}  \\  &
\textbf{\xspace COCOCO} & 21.34 & 0.66 & 
10.51 & 0.92 & 4.99 & 6.73 & 17.50 & 0.33 \\  &
\textbf{\xspace Cog-Inp} & 23.92 & 0.79 & 
10.78 & \textcolor{Blue}{\textbf{0.47}} & 3.23 & 7.03 & \textcolor{Blue}{\textbf{17.53}} & 0.17  \\  &
\textbf{\xspace Ours} & \textcolor{Red}{\textbf{25.27}}&\textcolor{Red}{\textbf{0.94}}&\textcolor{Red}{\textbf{4.29}}&\textcolor{Red}{\textbf{0.45}}&\textcolor{Red}{\textbf{1.41}}&\textcolor{Blue}{\textbf{7.21}}&\textcolor{Red}{\textbf{18.46}} &\textcolor{Red}{\textbf{0.09}}   \\
\bottomrule
\end{tabular}
}}
\label{tab:inp}
\end{table}

\subsection{Video Editing}

\OurMethod~can be used for video inpainting by employing Vison Language Models~\cite{gpt4,team2024gemini} to generate modified captions based on user editing instructions and source captions and apply \OurMethod~to inpaint based on the modified captions.
Tab.~\ref{tab:edit} shows the quantitative comparison on \Benchmark. 
We compare the editing results of inverse-based UniEdit~\cite{bai2024uniedit}, DiT-based DiTCtrl~\cite{cai2024ditctrl}, and end-to-end ReVideo~\cite{mou2024revideo}.
For both standard and long videos in \Benchmark, \OurMethod~achieves superior performance, even surpassing the end-to-end ReVideo. This success can be attributed to its dual-branch architecture, which ensures excellent background preservation and foreground generation capabilities, maintaining high fidelity in non-edited regions while ensuring edited regions closely align with editing instructions, complemented by inpainting region ID resampling that maintains ID consistency in long video.
%
%
The qualitative comparison with previous video inpainting methods is shown in Fig.~\ref{fig:inp}. 
\OurMethod~demonstrates superior performance in preserving visual fidelity and text-prompt consistency. 

\begin{table}[t]
\centering
\scriptsize
\caption{\textbf{Quantitative comparisons among \OurMethod~and other video editing models in \Benchmark~(Standard and Long Video)}: UniEdit~\cite{bai2024uniedit}, DitCtrl~\cite{cai2024ditctrl}, and ReVideo~\cite{mou2024revideo}. Metrics include masked region preservation, text alignment, and video quality. \textcolor{Red}{\textbf{Red}} stands for the best, \textcolor{Blue}{\textbf{Blue}} stands for the second best. }
\scalebox{0.85}{
\setlength{\tabcolsep}{0.2mm}{
\begin{tabular}{cl|ccccc|cc|c}
\toprule
\multicolumn{2}{c|}{\textbf{Metrics}}  & \multicolumn{5}{c|}{\textbf{Masked Region Preservation}} & \multicolumn{2}{c|}{\textbf{Text Alignment}} & \textbf{Video Quality} \\
\midrule
\multicolumn{2}{c|}{\textbf{Models}}    & \textbf{PSNR}$\uparrow$   & \textbf{SSIM}$\uparrow$  & \textbf{LPIPS}$_{^{\times 10^2}}$$\downarrow$   & \textbf{MSE}$_{^{\times 10^2}}$$\downarrow$     & \textbf{MAE}$_{^{\times 10^2}}$$\downarrow$  & \textbf{CLIP Sim}$\uparrow$      & \textbf{CLIP Sim (M)}$\uparrow$  & \textbf{FVID}$\downarrow$   \\ \midrule
\multirow{4}{*}{\rotatebox{90}{\textbf{Standard}}}  & 
\textbf{\xspace UniEdit} & 9.96 & 0.36 & 
56.68 & 11.08 & 25.78 & 8.46 & 14.23 & 1.36  \\  &
\textbf{\xspace DitCtrl} & 9.30 & 0.33 & 
57.42 & 12.73 & 27.45 & 8.52 & 15.59 & 0.57  \\  &
\textbf{\xspace ReVideo} & \textcolor{Blue}{\textbf{15.52}} & \textcolor{Blue}{\textbf{0.49}} & 
\textcolor{Blue}{\textbf{27.68}} & \textcolor{Blue}{\textbf{3.49}} & \textcolor{Blue}{\textbf{11.14}} & \textcolor{Red}{\textbf{9.34}} & \textcolor{Blue}{\textbf{20.01}} & \textcolor{Blue}{\textbf{0.42}}  \\  &
\textbf{\xspace Ours} & \textcolor{Red}{\textbf{22.63}}&\textcolor{Red}{\textbf{0.91}}&\textcolor{Red}{\textbf{7.65}}&\textcolor{Red}{\textbf{1.02}}&\textcolor{Red}{\textbf{2.90}}&\textcolor{Blue}{\textbf{8.67}}&\textcolor{Red}{\textbf{20.20}} & \textcolor{Red}{\textbf{0.18}} \\ 
\midrule
\multirow{4}{*}{\rotatebox{90}{\textbf{Long}}} &  
\textbf{\xspace UniEdit} & 10.37 & 0.30 & 
54.61 & 10.25 & 24.89 & 10.85 & 15.42 & 1.00  \\  &
\textbf{\xspace DitCtrl} & 9.76 & 0.28 & 
62.49 & 11.50 & 26.64 & \textcolor{Blue}{\textbf{11.78}} & 16.52 & 0.56  \\  &
\textbf{\xspace ReVideo} & \textcolor{Blue}{\textbf{15.50}} & \textcolor{Blue}{\textbf{0.46}} & 
\textcolor{Blue}{\textbf{28.57}} & \textcolor{Blue}{\textbf{3.92}} & \textcolor{Blue}{\textbf{12.24}} & 11.22 & \textcolor{Red}{\textbf{20.50}} & \textcolor{Blue}{\textbf{0.35}}  \\  &
\textbf{\xspace Ours} & \textcolor{Red}{\textbf{22.60}}&\textcolor{Red}{\textbf{0.90}}&\textcolor{Red}{\textbf{7.53}}&\textcolor{Red}{\textbf{0.86}}&\textcolor{Red}{\textbf{2.76}}&\textcolor{Red}{\textbf{11.85}}&\textcolor{Blue}{\textbf{19.38}} & \textcolor{Red}{\textbf{0.11}}   \\
\bottomrule
\end{tabular}
}}
\label{tab:edit}
\end{table}

\begin{figure*}[htbp]
  \centering
  \includegraphics[width=0.95\textwidth]{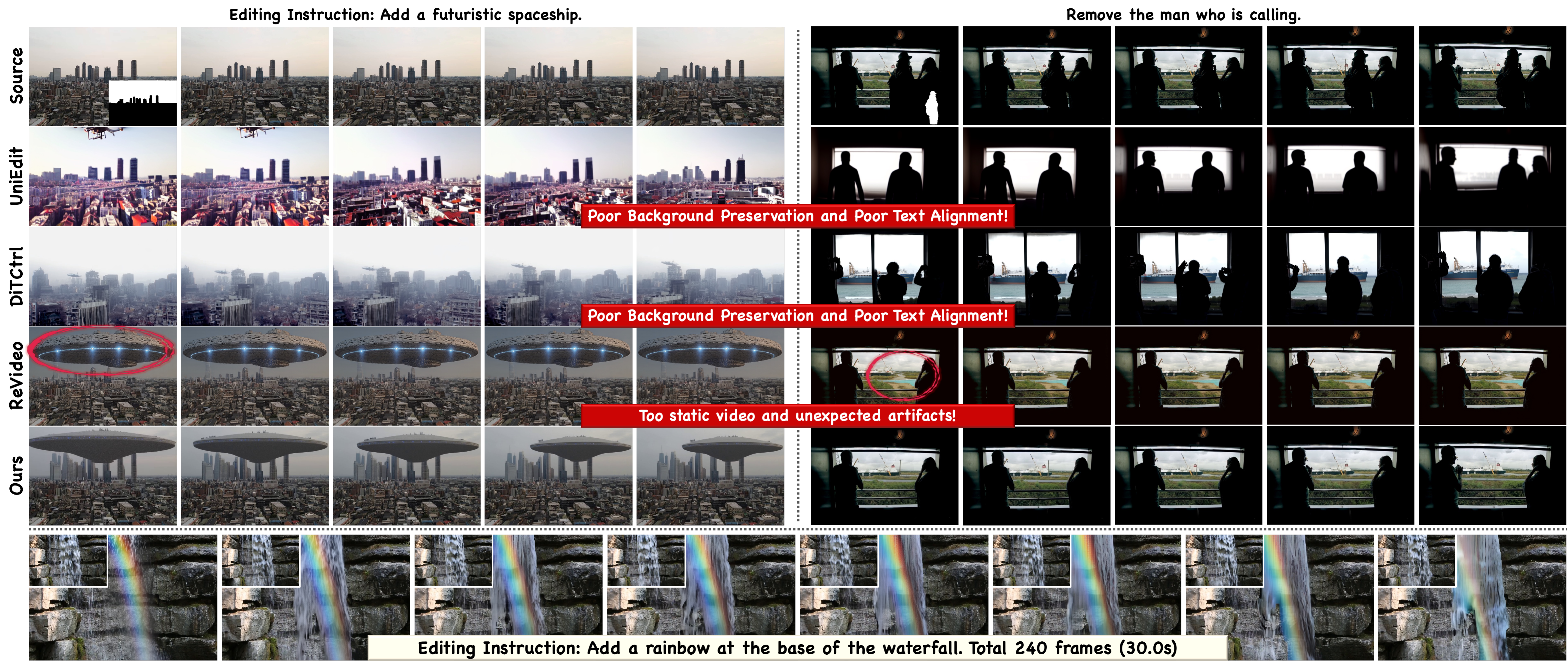} 
  \caption{Comparison of previous editing methods and VideoPainter on standard and long video editing.
  More visualizations are in the demo video.
  }
  \label{fig:edit}
\end{figure*}

\begin{table}[t]
\centering
\scriptsize
\caption{\textbf{User Study: User preference ratios comparing \OurMethod~with video inpainting and editing baselines.} For each sample, participants selected only one model that produced the best results for each criterion. We evaluate performance using the average proportion of being selected as the best response. For video inpainting, we compared \OurMethod~against ProPainter~\cite{Propainter}, COCOCO~\cite{zi2024cococo}, and Cog-Inp~\cite{yang2024cogvideox}. For video editing, we compared \OurMethod~against UniEdit~\cite{bai2024uniedit}, DitCtrl~\cite{cai2024ditctrl}, and ReVideo~\cite{mou2024revideo}. Detailed results are in the appendix.}
\scalebox{0.90}{
\setlength{\tabcolsep}{2mm}{
\begin{tabular}{l|ccc|ccc}
\toprule
\multirow{2}{*}{\textbf{Task}} & \multicolumn{3}{c|}{\textbf{Video Inpainting}} & \multicolumn{3}{c}{\textbf{Video Editing}} \\
\cmidrule{2-7}
& \textbf{Background} & \textbf{Text} & \textbf{Video} & \textbf{Background} & \textbf{Text} & \textbf{Video} \\
& \textbf{Preservation} & \textbf{Alignment} & \textbf{Quality} & \textbf{Preservation} & \textbf{Alignment} & \textbf{Quality} \\
\midrule
\textbf{Ours} & $74.2\%$ & $82.5\%$ & $87.4\%$ & $78.4\%$ & $76.1\%$ & $81.7\%$ \\
\bottomrule
\end{tabular}
}}
\label{tab:user_study}
\end{table}
\subsection{Human Evaluation}
We conducted a user study on video inpainting and editing tasks using standard-length video samples from the \Benchmark~inpainting and editing subsets. Thirty participants evaluated 50 randomly selected cases based on background preservation, text alignment, and video quality. As shown in Tab.~\ref{tab:user_study}, \OurMethod~significantly outperformed existing baselines, achieving higher preference rates across all evaluation criteria in both tasks. 
Detailed experiment settings and results are provided in the Appendix.

\subsection{Ablation Analysis}
We ablate on \OurMethod~in Tab .\ref{tab:ablation}, including architecture, context encoder size, control strategy, and inpainting region ID resampling.

\begin{table}[t]
\centering
\scriptsize
\caption{
\textbf{Ablation Studies on \Benchmark \hspace{-1pt}. }
\textbf{Single-Branch:} We add input channels to adapt masked video and finetune the backbone.
\textbf{Layer Configuration~(\OurMethod~(*)):} We vary the context encoder depth from one to four layers.
\textbf{w/o Selective Token Integration~(w/o Select):}: We bypass the token pre-selection step and integrate all context encoder tokens into DiT.
\textbf{T2V Backbone~(\OurMethod~(T2V)):} We replace the backbone from image-to-video DiTs to text-to-video DiTs.
\textbf{w/o target region ID resampling~(w/o Resample): }We ablate on the target region ID resampling.
(L) denotes evaluation on the long video subset.
\textcolor{Red}{\textbf{Red}} stands for the best result.}
\scalebox{0.82}{
\setlength{\tabcolsep}{0.2mm}{
\begin{tabular}{l|ccccc|cc|c}
\toprule
\multicolumn{1}{c|}{\textbf{Metrics}}  & \multicolumn{5}{c|}{\textbf{Masked Region Preservation}} & \multicolumn{2}{c|}{\textbf{Text Alignment}} & \textbf{Video Quality} \\
\midrule
\multicolumn{1}{c|}{\textbf{Models}}    & \textbf{PSNR}$\uparrow$   & \textbf{SSIM}$\uparrow$  & \textbf{LPIPS}$_{^{\times 10^2}}$$\downarrow$   & \textbf{MSE}$_{^{\times 10^2}}$$\downarrow$     & \textbf{MAE}$_{^{\times 10^2}}$$\downarrow$  & \textbf{CLIP Sim}$\uparrow$      & \textbf{CLIP Sim (M)}$\uparrow$  & \textbf{FVID}$\downarrow$   \\ \midrule
\textbf{\xspace Single-Branch} & 20.54 & 0.79 & 10.48 & 0.94 & 4.16 & 8.19 & 19.31 & 0.22  \\
\midrule
\textbf{\xspace \OurMethod~(1)} & 21.92 & 0.81 & 8.78 & 0.89 & 3.26 & 8.44 & 20.79 & 0.17  \\
\textbf{\xspace \OurMethod~(4)} & 22.86 & 0.85 & \textcolor{Red}{\textbf{6.51}} & 0.83 & 2.86 & 9.12 & 20.49 & 0.16  \\
\midrule
\textbf{\xspace w/o Select} & 20.94 & 0.74 & 7.90 & 0.95 & 3.87 & 8.26 & 17.84 & 0.25  \\
\midrule
\textbf{\xspace \OurMethod~(T2V)} & 23.01 & 0.87 & 6.94 & 0.89 & 2.65 & \textcolor{Red}{\textbf{9.41}} & 20.66 & 0.16  \\
\midrule
\textbf{\xspace \OurMethod} & \textcolor{Red}{\textbf{23.32}}&\textcolor{Red}{\textbf{0.89}}&6.85&\textcolor{Red}{\textbf{0.82}}&\textcolor{Red}{\textbf{2.62}}&8.66&\textcolor{Red}{\textbf{21.49}} & \textcolor{Red}{\textbf{0.15}} \\ 
\midrule
\midrule
\textbf{\xspace w/o Resample (L)} & 21.79 & 0.84 & \textcolor{Red}{\textbf{8.65}} & 0.81 & 3.10 & 11.35 & 20.68 & 0.19  \\
\midrule
\textbf{\xspace \OurMethod~(L)} & \textcolor{Red}{\textbf{22.19}}&\textcolor{Red}{\textbf{0.85}}&9.14&\textcolor{Red}{\textbf{0.71}}&\textcolor{Red}{\textbf{2.92}}&\textcolor{Red}{\textbf{11.52}}&\textcolor{Red}{\textbf{21.54}} &\textcolor{Red}{\textbf{0.17}}   \\
\bottomrule
\end{tabular}
}}
\label{tab:ablation}
\end{table}

Based on rows 1 and 5, the dual-branch \OurMethod~significantly outperforms its single-branch counterpart by explicitly decoupling background preservation from foreground generation, thereby reducing model complexity and avoiding the trade-off between competing objectives in a single branch.
Row 2 to row 6 of Tab. \ref{tab:ablation} demonstrate the rationale behind our key design choices: \blackcircle{1} utilizing a two-layer structure as an optimal balance between performance and efficiency for the context encoder, \blackcircle{2} implementing token-selective feature fusion based on segmentation mask information to prevent confusion from indistinguishable foreground-background tokens in the backbone, and \blackcircle{3} adapting plug-and-play control to different backbones with comparable performance. Furthermore, rows 7 and 8 verify the importance of employing inpainting region ID resampling for long videos, which maintains ID consistency by explicitly resampling inpainted region tokens from previous clips.

\subsection{Plug-and-Play Control Ability}
\begin{figure}[t]
  \centering
  \includegraphics[width=0.48\textwidth]{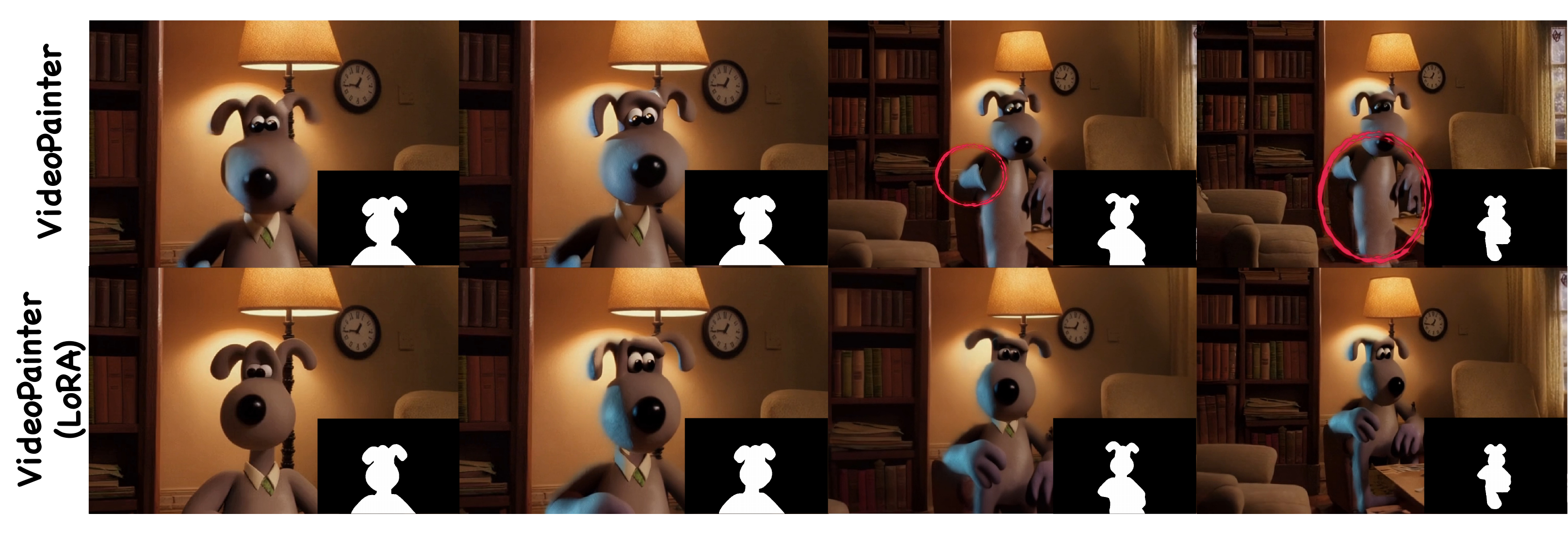} 
  \vspace{-24pt}
\caption{Integrating \OurMethod~to Gromit-style LoRA~\cite{CogVideoXLoRA2024}.}
  \vspace{-12pt}
  \label{fig:lora}
\end{figure}

Fig. \ref{fig:lora} demonstrates the flexible plug-and-play control capabilities of \OurMethod~in base diffusion transformer selection. We showcase how \OurMethod~can be seamlessly integrated with community-developed Gromit-style LoRA. Despite the significant domain gap between anime-style data and our training dataset, \OurMethod's dual-branch architecture ensures its plug-and-play inpainting abilities, enabling users to select the most appropriate base model for specific inpainting requirements and expected results.

\section{Discussion}\label{sec-5-con}

In this paper, we introduce \OurMethod, the first dual-branch video inpainting framework with plug-and-play control capabilities. Our approach features three key innovations: (1) a lightweight plug-and-play context encoder compatible with any pre-trained video DiTs, (2) an inpainting region ID resampling technique for maintaining long video ID consistency, and (3) a scalable dataset pipeline that produced \TrainingData~and \Benchmark, containing over 390K video clips with precise masks and dense captions. \OurMethod~also shows promise in video editing applications. Extensive experiments demonstrate that \OurMethod~achieves state-of-the-art performance across 8 metrics in video inpainting and editing, particularly in video quality, mask region preservation, and text coherence. 

However, \OurMethod~still has limitations: (1) Generation quality is limited by the base model, which may struggle with complex physical and motion modeling, and (2) performance is suboptimal with low-quality masks or misaligned video captions.

\section*{Acknowledgements}
This work was supported in part by the CUHK Strategic Seed Funding for Collaborative Research Scheme under Grant No. 3136023 and in part by the CUHK Research Matching Scheme under Grant No. 7106937, 8601130, and 8601440.

\newpage
\bibliographystyle{ACM-Reference-Format}
\bibliography{sample-bibliography}

\newpage
\appendix

\section{User Study}

For a comprehensive evaluation beyond the quantitative and qualitative experiments presented in sections 4.2 and 4.3, we conducted a user study on both video inpainting and editing tasks using standard-length video samples from the \Benchmark~inpainting subset and editing subset. The study involved 30 participants evaluating 50 randomly selected cases across three criteria: background preservation, text alignment, and video quality. 
Specifically, we assessed (1) Background Preservation: the error distance between the uninpainted regions of the generated video and the original video, (2) Text Alignment: the semantic coherence between inpainted regions and text captions, and (3) Video Quality: the overall visual fidelity of the generated results. As shown in Tab.~\ref{tab:user_study_full}, our approach demonstrates superior performance across all three aspects in both tasks. For video inpainting, our method achieves win rates of \textbf{74.2\%}, \textbf{82.5\%}, and \textbf{87.4\%} respectively. Similarly, for video editing, we obtain win rates of \textbf{78.4\%}, \textbf{76.1\%}, and \textbf{81.7\%} across the three aspects.

\begin{table}[htbp]
\centering
\small
\caption{\textbf{User study evaluation comparing \OurMethod~against state-of-the-art video inpainting and editing models}. We conducted comprehensive comparisons on the \Benchmark, randomly sampling 50 examples from each of the inpainting and editing subsets. Human evaluators assessed the models' outputs based on three criteria: background preservation, text alignment, and overall video quality. For each sample, participants selected only one model that produced the best results for each criterion. We evaluate performance using the proportion of model-generated outputs selected as the optimal response across all samples. For video inpainting, we compared against ProPainter~\cite{Propainter}, COCOCO~\cite{zi2024cococo}, and Cog-Inp~\cite{yang2024cogvideox}. For video editing, we compared against UniEdit~\cite{bai2024uniedit}, DitCtrl~\cite{cai2024ditctrl}, and ReVideo~\cite{mou2024revideo}. \textcolor{Red}{\textbf{Red}} stands for the best, \textcolor{Blue}{\textbf{Blue}} stands for the second best.}
\scalebox{0.9}{
\setlength{\tabcolsep}{1.5mm}{
\begin{tabular}{cl|ccc}
\toprule
\multicolumn{2}{c|}{\multirow{2}{*}{\textbf{Metrics}}} & \textbf{Background} & \textbf{Text} & \textbf{Video} \\
& & \textbf{Preservation}$(\%)\uparrow$ & \textbf{Alignment}$(\%)\uparrow$ & \textbf{Quality}$(\%)\uparrow$ \\
\midrule
\multirow{4}{*}{\rotatebox{90}{\textbf{Inpainting}}} & 
\textbf{\xspace ProPainter} & 3.5 & 2.5 & 1.7 \\  &
\textbf{\xspace COCOCO} & 7.0 & 2.7 & 2.1 \\  &
\textbf{\xspace Cog-Inp} & \textcolor{Blue}{\textbf{15.3}} & \textcolor{Blue}{\textbf{12.3}} & \textcolor{Blue}{\textbf{8.8}} \\  &
\textbf{\xspace Ours} & \textcolor{Red}{\textbf{74.2}} & \textcolor{Red}{\textbf{82.5}} & \textcolor{Red}{\textbf{87.4}} \\ 
\midrule
\multirow{4}{*}{\rotatebox{90}{\textbf{Editing}}} & 
\textbf{\xspace UniEdit} & 3.3 & 3.7 & 2.8 \\  &
\textbf{\xspace DitCtrl} & 5.8 & 6.1 & 4.2 \\  &
\textbf{\xspace ReVideo} & \textcolor{Blue}{\textbf{12.5}} & \textcolor{Blue}{\textbf{14.1}} & \textcolor{Blue}{\textbf{11.3}} \\  &
\textbf{\xspace Ours} & \textcolor{Red}{\textbf{78.4}} & \textcolor{Red}{\textbf{76.1}} & \textcolor{Red}{\textbf{81.7}} \\
\bottomrule
\end{tabular}
}}
\label{tab:user_study_full}
\end{table}

\section{Ablation on Dual-Branch Artecture}

\begin{figure}[t]
  \centering
  \includegraphics[width=0.48\textwidth]{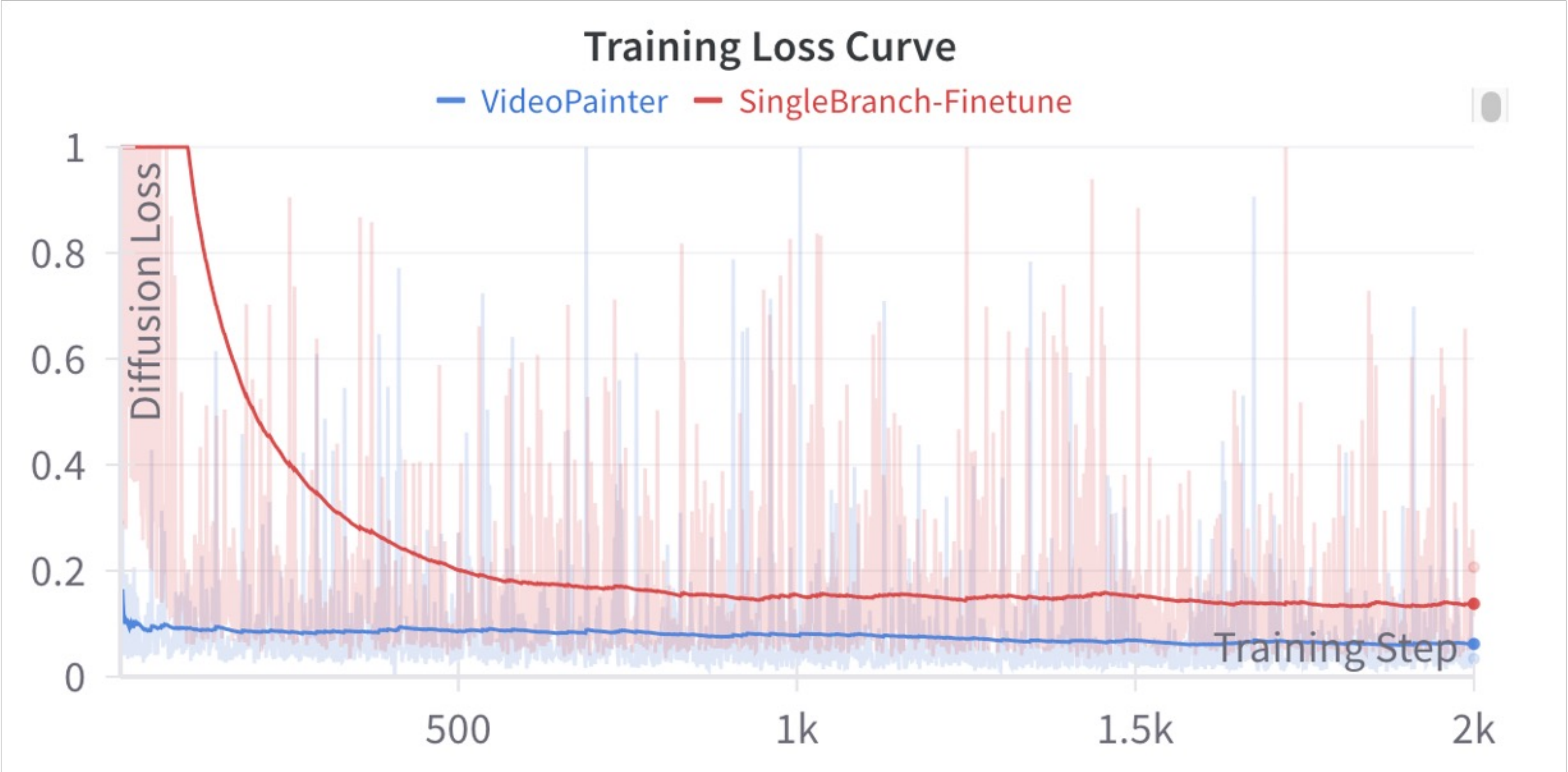} 
  \caption{Training loss curve of ablation of single branch fine-tuning and default dual branch \OurMethod. The training loss curves demonstrate that our dual-branch \OurMethod~achieves superior convergence speed, stability, and final performance compared to single-branch fine-tuning, despite having significantly fewer trainable parameters.}
  \vspace{-10pt}
  \label{fig:curve}
\end{figure}

As shown in Tab. \ref{tab:ablation_on_branch} and Fig. \ref{fig:curve}, we ablate on \OurMethod~with a single branch fine-tuning setup.

Tab. \ref{tab:ablation_on_branch} clearly shows that the dual-branch \OurMethod~significantly outperforms its single-branch counterpart by explicitly decoupling background preservation from foreground generation, thereby reducing model complexity and avoiding the trade-off between competing objectives in a single branch. 
The training loss curves in Fig. \ref{fig:curve} demonstrate that \OurMethod, through effective dual-branch decoupling, achieves an advantageous optimization starting point and maintains stable, efficient convergence. By separating the competing objectives of background preservation and foreground generation via follow caption into distinct branches rather than a single backbone, \OurMethod~achieves superior convergence performance, with a final loss approximately half that of single-branch finetuning.

\section{Ablation on Mask Quality}
\begin{table}[t]
\centering
\scriptsize
\caption{
\textbf{Ablation Studies on \Benchmark}:
Kernel (*): We randomly sample dilation and erosion with varying kernel sizes $\in (8, 16, 32)$ for the segmentation masks. Kernel (Square): We randomly sample square masks with varying sizes $\in [8, 32]$ and random locations. This reflects \OurMethod's robustness to different mask qualities.
\textcolor{Red}{\textbf{Red}} stands for the best result.
}
\scalebox{0.82}{
\setlength{\tabcolsep}{0.2mm}{
\begin{tabular}{l|ccccc|cc|c}
\toprule
\multicolumn{1}{c|}{\textbf{Metrics}}  & \multicolumn{5}{c|}{\textbf{Masked Region Preservation}} & \multicolumn{2}{c|}{\textbf{Text Alignment}} & \textbf{Video Quality} \\
\midrule
\multicolumn{1}{c|}{\textbf{Models}}    & \textbf{PSNR}$\uparrow$   & \textbf{SSIM}$\uparrow$  & \textbf{LPIPS}$_{^{\times 10^2}}$$\downarrow$   & \textbf{MSE}$_{^{\times 10^2}}$$\downarrow$     & \textbf{MAE}$_{^{\times 10^2}}$$\downarrow$  & \textbf{CLIP Sim}$\uparrow$      & \textbf{CLIP Sim (M)}$\uparrow$  & \textbf{FVID}$\downarrow$   \\ \midrule
\textbf{\xspace Kernel (8)} & 23.24 & 0.84 & 7.43 & 0.84 & 2.69 & 8.51 & 21.18 & 0.15  \\
\textbf{\xspace Kernel (16)} & 23.17 & 0.82 & 7.55 & 0.87 & 2.93 & 8.49 & 20.37 & 0.17  \\
\textbf{\xspace Kernel (32)} & 23.02 & 0.81 & 7.64 & 0.92 & 3.07 & 8.28 & 20.38 & 0.16  \\
\textbf{\xspace Kernel (Square)} & 22.46 & 0.73 & 9.37 & 0.94 & 3.75 & 7.81 & 19.88 & 0.21  \\
\midrule

\textbf{\xspace \OurMethod} & \textcolor{Red}{\textbf{23.32}}&\textcolor{Red}{\textbf{0.89}}&\textcolor{Red}{\textbf{6.85}}&\textcolor{Red}{\textbf{0.82}}&\textcolor{Red}{\textbf{2.62}}&\textcolor{Red}{\textbf{8.66}}&\textcolor{Red}{\textbf{21.49}} & \textcolor{Red}{\textbf{0.15}} \\ 

\bottomrule
\end{tabular}
}}
\label{tab:ablation_on_mask_quality}
\end{table}

As shown in Tab. \ref{tab:ablation_on_mask_quality}, we perform ablation studies on \OurMethod~using a single branch fine-tuning configuration.

Tab. \ref{tab:ablation_on_mask_quality} clearly demonstrates that \OurMethod~exhibits robust generalization across various qualities of video segmentation masks (which we augmented by randomly sampling different kernel sizes for dilation and erosion operations applied to the original segmentation masks). However, when utilizing random rectangular masks (which are highly atypical in video segmentation), model performance decreased by approximately 13\%. This performance degradation primarily stems from artifacts generated along the rectangular edges, which we attribute to the substantial geometric disparity between these artificial rectangular masks and natural segmentation masks encountered in real-world scenarios.

\section{Ablation on ID Resampling}
\begin{table}[t]
\centering
\scriptsize
\caption{
\textbf{Ablation Studies on \Benchmark}:
w/o target region ID resampling~(w/o Resample): We ablate on the target region ID resampling. (L) denotes evaluation on the long video subset. 
\textcolor{Red}{\textbf{Red}} stands for the best result.}
\scalebox{0.82}{
\setlength{\tabcolsep}{0.2mm}{
\begin{tabular}{l|ccccc|cc|c}
\toprule
\multicolumn{1}{c|}{\textbf{Metrics}}  & \multicolumn{5}{c|}{\textbf{Masked Region Preservation}} & \multicolumn{2}{c|}{\textbf{Text Alignment}} & \textbf{Video Quality} \\
\midrule
\multicolumn{1}{c|}{\textbf{Models}}    & \textbf{PSNR}$\uparrow$   & \textbf{SSIM}$\uparrow$  & \textbf{LPIPS}$_{^{\times 10^2}}$$\downarrow$   & \textbf{MSE}$_{^{\times 10^2}}$$\downarrow$     & \textbf{MAE}$_{^{\times 10^2}}$$\downarrow$  & \textbf{CLIP Sim}$\uparrow$      & \textbf{CLIP Sim (M)}$\uparrow$  & \textbf{FVID}$\downarrow$   \\ \midrule
\textbf{\xspace w/o Resample (L)} & 21.79 & 0.84 & \textcolor{Red}{\textbf{8.65}} & 0.81 & 3.10 & 11.35 & 20.68 & 0.19  \\
\midrule
\textbf{\xspace \OurMethod~(L)} & \textcolor{Red}{\textbf{22.19}}&\textcolor{Red}{\textbf{0.85}}&9.14&\textcolor{Red}{\textbf{0.71}}&\textcolor{Red}{\textbf{2.92}}&\textcolor{Red}{\textbf{11.52}}&\textcolor{Red}{\textbf{21.54}} &\textcolor{Red}{\textbf{0.17}}   \\
\bottomrule
\end{tabular}
}}
\label{tab:ablation_on_id_resampling}
\end{table}
\begin{figure*}[t]
  \centering
  \includegraphics[width=0.98\textwidth]{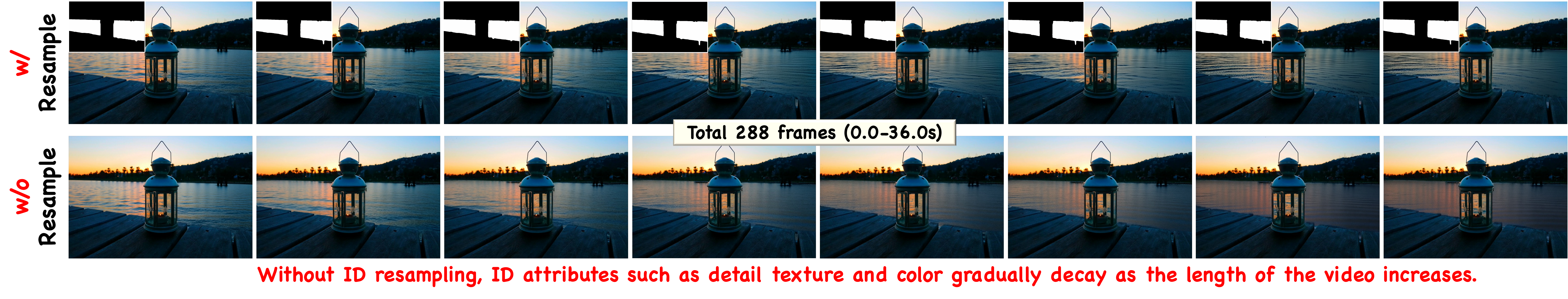} 
  \caption{
  \textcolor{red}{Long video inpainting comparison of original VideoPainter and w/o ID resampling ablation version.}
  }
  \label{fig:ablation_on_id_resampling}
\end{figure*}

As shown in Tab. \ref{tab:ablation_on_id_resampling} and Fig. \ref{fig:ablation_on_id_resampling}, we ablate on \OurMethod~with a w/o ID resampling ablation version.

Tab. \ref{tab:ablation_on_branch} clearly verifies the importance of employing inpainting region ID resampling for long videos, which maintains ID consistency by explicitly resampling inpainted region tokens from previous clips.
Fig. \ref{fig:curve} demonstrates that removing the ID resample adapter degrades long-video inpainting performance, with more severe deterioration as video length increases.

\section{More Related Works}

\subsection{Video Inpainting}
Video inpainting approaches can be broadly classified into two categories based on whether they possess generative capabilities:

\paragraph{Non-generative methods.} 
These methods~\cite{Propainter,EEFVI,3dcnn3,transformer4,optical7} leverage architecture priors to facilitate pixel propagation. This includes utilizing local perception of 3D CNNs~\cite{3dcnn1,3dcnn2,3dcnn3,temporal_shift}, and exploiting the global perception of attention to retrieve and aggregate tokens with similar texture for filling masked video~\cite{transformer1,transformer2,transformer3,transformer4}.
They also introduce various physical quantities, especially optical flow, as auxiliary conditions as it simplifies RGB pixel inpainting by completing less complex flow fields~\cite{optical1,optical2,optical3,optical4,optical5,optical6,optical7}. 
A notable exemplar of this approach is ProPainter~\cite{Propainter}, which integrates flow completion models with spatiotemporal transformers into an end-to-end framework, achieving more faithful propagation through combined pixel and feature propagation with flow consistency verification. 
However, they are only effective for partial object occlusions with random masks but face significant limitations when inpaint fully masked regions due to insufficient contexts.

\paragraph{Generative methods.}
Recent advances in generative foundation models~\cite{rombach2022high,guo2023animatediff} have sparked numerous approaches that leverage additional modules or training strategies to extend backbones' capabilities for video inpainting~\cite{zhang2024avid,zi2024cococo,wang2024videocomposer}.
VideoComposer~\cite{wang2024videocomposer}, built upon a pre-trained video generation backbone Stable Video Diffusion~\cite{blattmann2023stable}, integrates various control signals (text, depth, mask, motion vectors) through a shared spatiotemporal condition fusion module. While it enables limited inpainting capabilities through mask and text conditioning, its performance in background preservation is compromised due to control condition compression and lack of masked video conditioning support.
AVID~\cite{zhang2024avid} and COCOCO~\cite{zi2024cococo} represent the most related recent works. Both adopt a similar implementation by augmenting Stable Diffusion Inpainting~\cite{sd2inpainting} with trainable temporal attention layers. This architecture includes per-frame region filling based on the image inpainting backbone and temporal smoothing with temporal attention. 
Despite showing promising results for both random and segmentation masks due to their generative abilities, they struggle to balance background preservation and foreground generation with text caption~\cite{ju2024brushnet,li2024brushedit,bian2024motioncraft} within the single backbone.
AVID also explores any-length video inpainting by smoothing latent at segment boundaries and using the middle frame as the ID reference. 
However, identity discontinuities occur during significant variations, and identity degradation emerges in longer video generation scenarios.
In contrast, \OurMethod~is a dual-branch framework by decoupling video inpainting into foreground generation and background-guided preservation. It employs an efficient context encoder to guide any pre-trained DiT, facilitating plug-and-play control and zero-shot adaptation across various stylization backbones—a capability absent in both non-generative and generative methods. 
Furthermore, \OurMethod~also introduces a novel target region ID resampling technique that enables ID consistency in any-length video inpainting by explicitly resampling the inpainted region of the last generated clip.

\begin{table}[t]
\centering
\caption{Comparison of video inpainting datasets. Our \TrainingData~is the largest video inpainting dataset to date, comprising over 390K
high-quality clips with segmentation masks, video captions, and masked region descriptions.}
\vspace{-6pt}
\label{tab-dataset_comparison}
\setlength{\tabcolsep}{0.9mm}
\scalebox{0.71}{
\begin{tabular}{l|c|c|c|c}
\toprule
Dataset & \#Clips & Duration & Video Caption & Masked Region Desc. \\
\midrule
DAVIS~\cite{perazzi2016benchmark} & 0.4K & 0.1h & \xmark & \xmark \\
YouTube-VOS~\cite{xu2018youtube} & 4.5K & 5.6h & \xmark & \xmark \\
VOST~\cite{tokmakov2023breaking} & 1.5K & 4.2h & \xmark & \xmark \\
MOSE~\cite{ding2023mose} & 5.2K & 7.4h & \xmark & \xmark \\
LVOS~\cite{hong2023lvos} & 1.0K & 18.9h & \xmark & \xmark \\
SA-V~\cite{ravi2024sam} & 642.6K & 196.0h & \xmark & \xmark \\
\bottomrule
Ours & 390.3K & 866.7h & \checkmark & \checkmark \\
\bottomrule
\end{tabular}
}
\vspace{-0.4cm}
\end{table}

\subsection{Video Inpainting Datasets}

Recent advances in segmentation~\cite{ravi2024sam} have created many video segmentation datasets~\cite{perazzi2016benchmark,xu2018youtube,hong2023lvos,ding2023mose,tokmakov2023breaking,VISOR2022}. Among these, DAVIS~\cite{perazzi2016benchmark} and YouTube-VOS~\cite{xu2018youtube} have become prominent benchmarks for video inpainting due to their high-quality masks and diverse object categories. However, the existing datasets face two primary limitations: (1) insufficient scale to meet the data requirements of generative models, and (2) the absence of crucial control conditions necessary for generating masked objects such as video captions. 
These constraints have impeded the evaluation of state-of-the-art generative approaches, with methods like COCOCO and AVID resorting to random sampling from annotated internet video datasets for assessment. Consequently, these limitations have also hindered the further advancement of generative approaches in video inpainting.
In contrast, as shown in Tab. \ref{tab-dataset_comparison}, we developed a scalable dataset pipeline based on state-of-the-art vision understanding models~\cite{zhang2024recognize,ravi2024sam,gpt4}, and constructed the largest video inpainting dataset to date with over 390K clips, each annotated with segmentation masks and dense video captions.
In contrast, we developed a scalable dataset construction pipeline based on state-of-the-art vision understanding models~\cite{zhang2024recognize,ravi2024sam,gpt4}, including open-set object detection, object segmentation mask tracking, shot transition detection, motion score and aesthetic quality filtering, and video annotation. By leveraging publicly available internet videos, we constructed the largest video inpainting dataset to date, comprising over 390K high-quality clips, each accompanied by precise segmentation masks, comprehensive video captions, and masked region descriptions, as Tab. \ref{tab-dataset_comparison} shows.

\section{More Discussions}

As mentioned in the main paper, VideoPainter's performance is suboptimal with low-quality or misaligned video captions. 
However, it should be noted that VideoPainter's text-following capability (including complex physical or motion modeling abilities) derives from pre-trained DiTs. This can be enhanced by substituting more powerful DiT models, as our method is a plug-and-play framework independent of the internal DiT backbone architecture.

\section{More Visualizations}

\begin{figure*}[!htbp]
  \centering
  \includegraphics[width=0.98\textwidth]{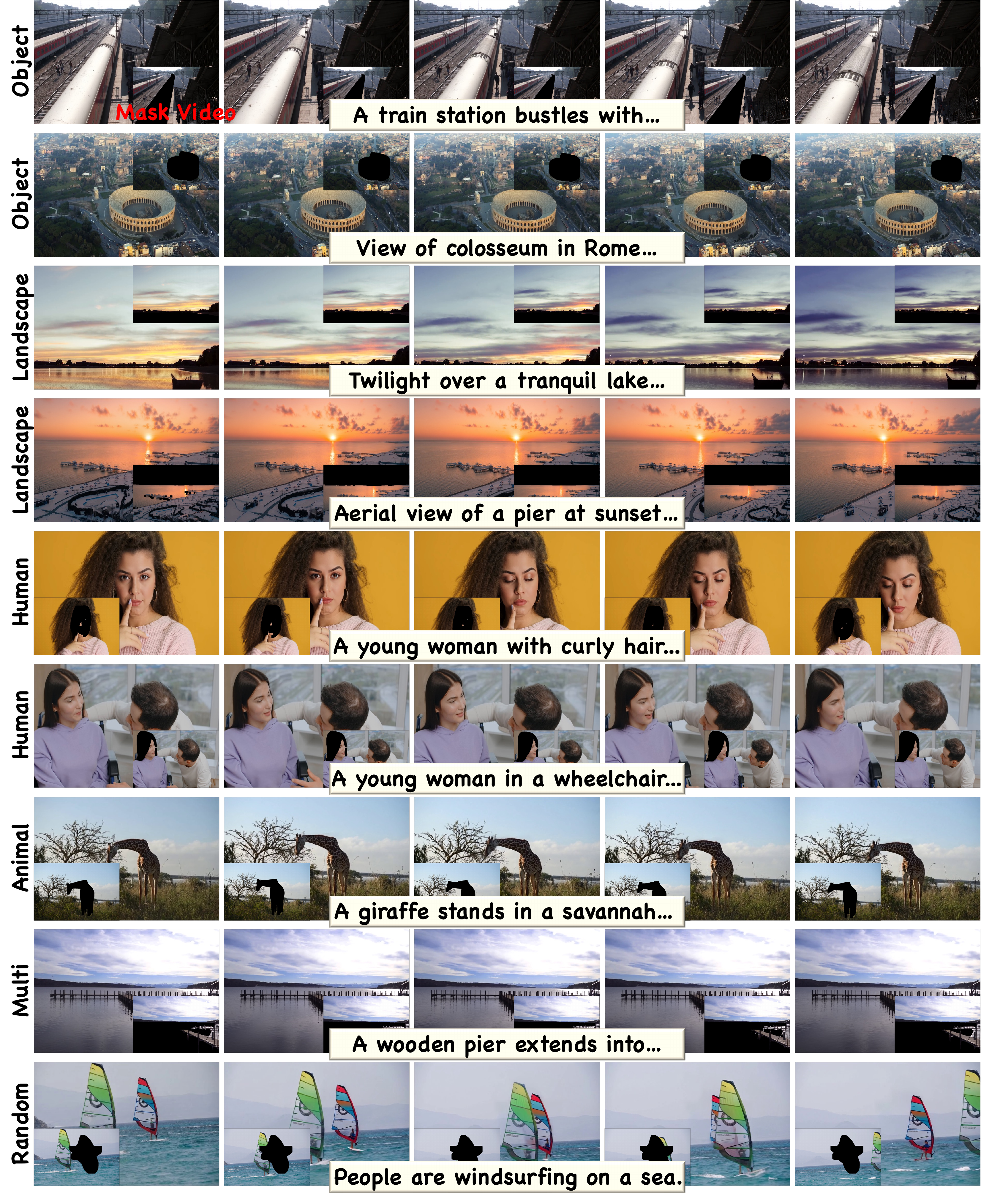} 
  \caption{More video inpainting results.}
  \label{fig:edit}
\end{figure*}

\begin{figure*}[!htbp]
  \centering
  \includegraphics[width=0.98\textwidth]{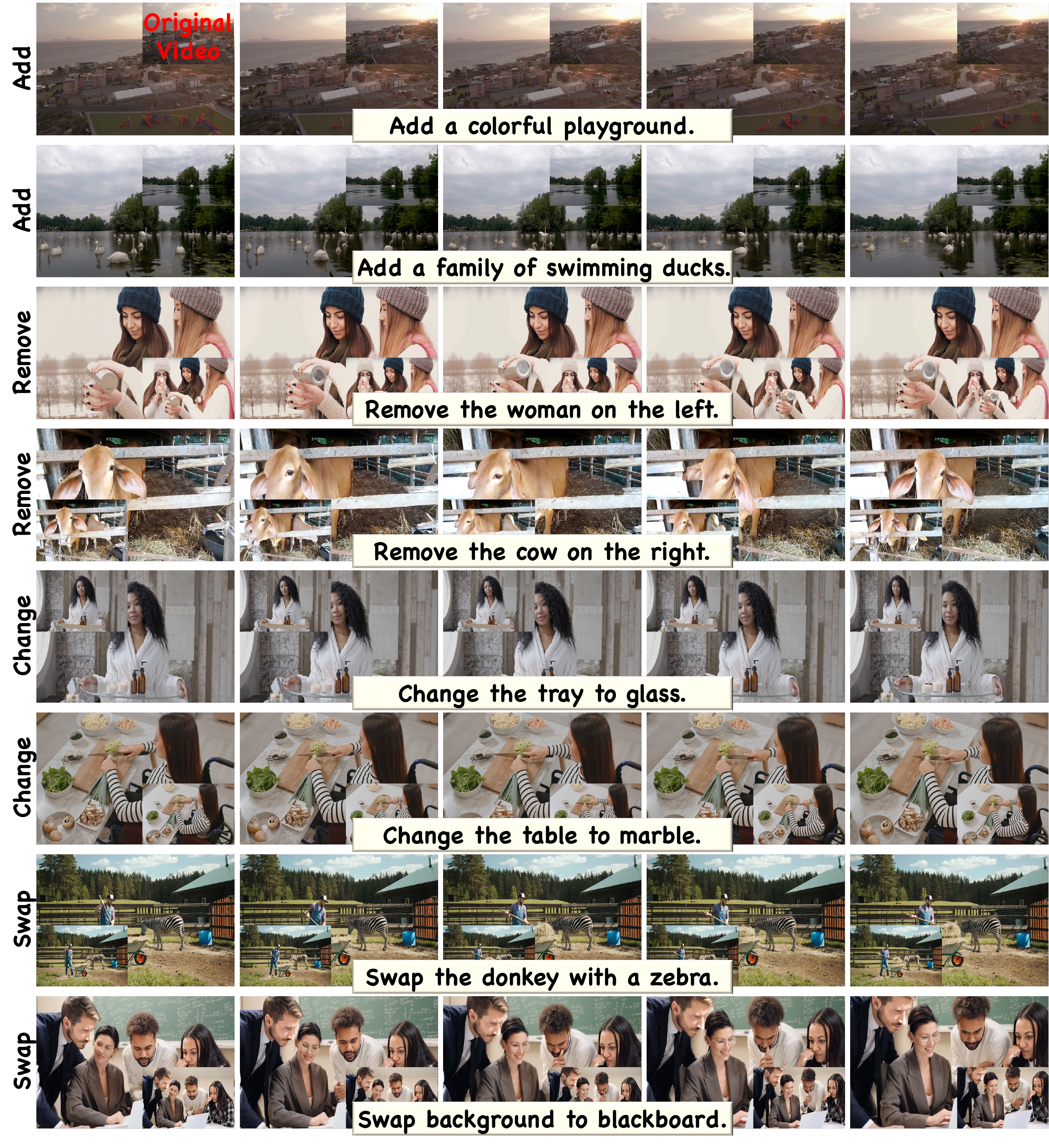} 
  \caption{More video editing results.}
  \label{fig:edit}
\end{figure*}

\end{document}